# Population Empirical Bayes

**Alp Kucukelbir**
Data Science Institute & Computer Science
Columbia University
New York, NY 10027

**David M. Blei**
Data Science Institute & Computer Science & Statistics
Columbia University
New York, NY 10027

## Abstract

Bayesian predictive inference analyzes a dataset to make predictions about new observations. When a model does not match the data, predictive accuracy suffers. We develop population empirical Bayes (POP-EB), a hierarchical framework that explicitly models the empirical population distribution as part of Bayesian analysis. We introduce a new concept, the latent dataset, as a hierarchical variable and set the empirical population as its prior. This leads to a new predictive density that mitigates model mismatch. We efficiently apply this method to complex models by proposing a stochastic variational inference algorithm, called bumping variational inference (BUMP-VI). We demonstrate improved predictive accuracy over classical Bayesian inference in three models: a linear regression model of health data, a Bayesian mixture model of natural images, and a latent Dirichlet allocation topic model of scientific documents.

## 1 INTRODUCTION

Bayesian modeling is a powerful framework for analyzing structured data. It covers many important methods in probabilistic machine learning, such as regression, mixture models, hidden Markov models, and probabilistic topic models (Bishop, 2006; Murphy, 2012). Bayesian models provide an intuitive language to express assumptions about data, as well as general-purpose algorithms (such as variational methods) to reason under those assumptions.

Bayesian models describe data $\mathbf{X}$ as a structured probability density with latent variables $\boldsymbol{\theta}$, $p(\mathbf{X}, \boldsymbol{\theta}) = p(\mathbf{X} \mid \boldsymbol{\theta}) p(\boldsymbol{\theta})$. The first term is the likelihood, the second is the prior. We use this joint density to both analyze data and form predictions. We analyze data through the posterior density of the latent variables, $p(\boldsymbol{\theta} \mid \mathbf{X})$. This conditional density derives from the joint. We form predictions by combining the likelihood and posterior density as

$$p(\boldsymbol{x}_{\text{new}} \mid \mathbf{X}) = \int p(\boldsymbol{x}_{\text{new}} \mid \boldsymbol{\theta}) p(\boldsymbol{\theta} \mid \mathbf{X}) \, \mathrm{d}\boldsymbol{\theta}. \qquad (1)$$

This is the Bayesian predictive density, the conditional density of a new observation given the dataset.

Frequentist statistics take a different perspective. Here we analyze a set of observations $\mathbf{X}$ to draw conclusions about the mechanism $F$ that gave rise to them. $F(\mathbf{X})$ is an unknown distribution; it is called the *population*. Nonparametric frequentist methods, like the bootstrap (Efron and Tibshirani, 1994), work directly with $F$ while still respecting its unknown nature. This leads to powerful tools that work well across many statistical settings. One goal is to use a given dataset to learn about $F$ and predict new observations; this is predictive inference (Young and Smith, 2005).

**Main idea.** We combine the flexibility of Bayesian modeling with the robustness of nonparametric frequentist statistics. The issue is that Bayesian theory, under frequentist scrutiny, assumes the model is correct (Bernardo and Smith, 2000). But this is rarely true; the model is almost always mismatched, which can lead to brittle data analysis and poor predictions. Our goal is to use the unknown population $F$ to improve a Bayesian model's predictive performance.

We call our framework population empirical Bayes (POP-EB). It describes a simple procedure. The input is a model $p(\mathbf{X}, \boldsymbol{\theta})$ and a dataset $\mathbf{X}$ of $N$ observations. For example, a mixture of Gaussians and a dataset of natural images.

1. Draw $B$ bootstrap samples of the dataset, $\{\mathbf{X}^{(1)}, \cdots, \mathbf{X}^{(B)}\}$. Each sample is a dataset of size $N$, drawn with replacement from the original dataset (Efron and Tibshirani, 1994).

2. Compute the posterior for each bootstrapped dataset, $p(\boldsymbol{\theta} \mid \mathbf{X}^{(b)})$. Evaluate the average predictive accuracy of the original dataset with Equation (1).

3. Pick the bootstrapped dataset $\mathbf{X}^{(b^*)}$ that best predicts the original dataset. Use the corresponding predictive density $p(\boldsymbol{x}_{\text{new}} \mid \mathbf{X}^{(b^*)})$ to form future predictions.

POP-EB applies to any Bayesian model and can improve its predictive performance. Above we describe its simplest form. Alternatives, which we discuss below, include one that weights the bootstrapped datasets and another that embeds the population into a stochastic variational inference algorithm (Hoffman et al., 2013).

In this paper, we develop, motivate, and study POP-EB. We show that it yields a predictive density that incorporates the unknown population distribution $F$ in a type of empirical Bayes model (Robbins, 1955, 1964). Compared to traditional Bayesian inference, it better predicts held-out data for models such as regression (Table 1), mixtures of Gaussians (Figure 5), and probabilistic topic models (Figure 7).

**Related work.** There are several running themes in this paper. The first is Bayesian/frequentist compromise and empirical Bayes. A rich literature relates Bayesian and frequentist ideas. Bayarri and Berger (2004) and Aitkin (2010) give thorough reviews. One thread of ideas around combining these two schools of thought is empirical Bayes (EB) (Robbins, 1955; Morris, 1983; Carlin and Louis, 2000). Loosely, EB uses frequentist statistics to estimate prior specifications in Bayesian models. EB enjoys good statistical properties (Efron, 2010); it can explain challenging concepts, such as the James-Stein paradox (Berger, 1985).

The second theme is predictive inference. One approach to Bayesian inference is to study the Bayesian predictive density. The goal is to design models such that the predictive density is high for new observations (Geisser, 1993). Machine learning also strives to develop models that deliver high predictive accuracy (Bishop, 2006). Our method explicitly optimizes the Bayesian predictive density.

A final theme is model misspecification. To paraphrase Box and Draper (1987), the challenge of Bayesian statistics is that while many models are useful, all of them are wrong. Robust statistics offer some remedies (Berger, 1994), such as using likelihoods and priors that are "insensitive to small deviations from the assumptions" (Huber and Ronchetti, 2009). Our work uses empirical Bayes to induce a model that is robust to misspecification.

## 2 POPULATION EMPIRICAL BAYES

Population empirical Bayes (POP-EB) incorporates the model-independent population $F$ into Bayesian analysis. We first develop the structure of our framework and then discuss the motivation behind it in Section 2.3.

### 2.1 EMPIRICAL BAYES

Let $\mathbf{X} = \{x_n\}_1^N$ be a dataset with $N$ observations. The dataset is a sample from an unknown population distribution $F$. The population is the "true" distribution of the data; it is independent of any model (Shao, 2003).

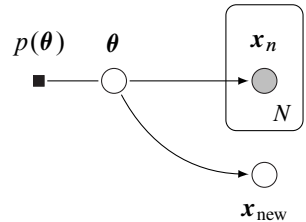

**Figure 1:** Graphical model for Bayesian predictive inference. The likelihood relates the dataset $\mathbf{X} = \{x_n\}_1^N$, along with the new observation $x_{\text{new}}$, to the latent variables $\boldsymbol{\theta}$. Conditioning on the observations and marginalizing over $\boldsymbol{\theta}$ gives the Bayesian predictive density of Equation (1).

A Bayesian model has two parts. The first is the likelihood, $p(x_n \mid \boldsymbol{\theta})$. It relates an observation $x_n$ to a set of latent random variables $\boldsymbol{\theta}$. If the observations are independent and identically distributed, the likelihood of the dataset becomes $p(\mathbf{X} \mid \boldsymbol{\theta}) = \prod_{n=1}^N p(x_n \mid \boldsymbol{\theta})$.

The second is the prior density, $p(\boldsymbol{\theta})$. This induces the joint density $p(\mathbf{X}, \boldsymbol{\theta}) = p(\mathbf{X} \mid \boldsymbol{\theta}) \, p(\boldsymbol{\theta})$. In predictive inference, we additionally consider a new observation $x_{\text{new}}$. It shares the same likelihood of the data, $p(x_{\text{new}} \mid \boldsymbol{\theta})$. This expands the joint density to $p(x_{\text{new}}, \mathbf{X}, \boldsymbol{\theta})$, shown in Figure 1.

A predictive density describes $x_{\text{new}}$ given the observed dataset $\mathbf{X}$. A simple recipe supplies such a density: condition on the observations and marginalize over the latent variables. This gives the Bayesian predictive density in Equation (1). It depends on the Bayesian posterior density,

$$p(\boldsymbol{\theta} \mid \mathbf{X}) = \frac{p(\mathbf{X} \mid \boldsymbol{\theta}) \, p(\boldsymbol{\theta})}{\int p(\mathbf{X} \mid \boldsymbol{\theta}') \, p(\boldsymbol{\theta}') \, \mathrm{d}\boldsymbol{\theta}'}, \qquad (2)$$

which describes how the latent variables $\boldsymbol{\theta}$ vary conditioned on a given dataset $\mathbf{X}$.

The main idea behind Empirical Bayes (EB) is to build estimates from the population into Bayesian inference (Robbins, 1955; Morris, 1983). EB uses the observed data to estimate parts of a Bayesian model. There are many variants of EB. Robbins (1955) proposed the following approach.

First, augment the model such that each observation $x_n$ gets its own set of latent variables $\boldsymbol{\theta}_n$. This is sometimes called the "compound sampling model" (Berger, 1985). It has connections to robust inference (Berger and Berliner, 1986) through an intuitive justification.[1] Then, assume the latent variables are exchangeable and distributed according to an unknown prior density $g$. Figure 2 shows the EB framework.

---
[1] When pondering outliers in Bayesian inference, de Finetti explains: "We know that each observation is taken using an instrument with [some] error, but each time chosen at random from a collection of instruments of different precisions" (de Finetti, 1961)

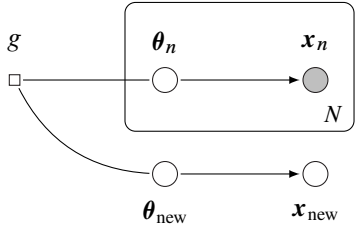

**Figure 2:** Graphical model for empirical Bayes (EB) predictive inference. We augment the Bayesian model and estimate the prior $g$ from the dataset.

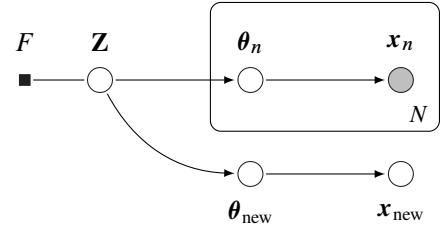

**Figure 3:** Graphical model for the population empirical Bayes (POP-EB) framework. We introduce $\mathbf{Z}$, the latent dataset, and assign the population distribution $F$ as its prior.

"Nonparametric EB" estimates the prior $g$ using nonparametric statistics of the data (Robbins, 1955, 1964). The name emphasizes its model-independent approach. "Parametric EB" assumes a parametric family for $g$ and estimates its parameters from the data (Morris, 1983). The result is a hierarchical Bayesian model whose top-level parameters are determined by the observations. The name emphasizes its model-based approach.

Both variants of EB introduce population information from the dataset $\mathbf{X}$. But neither directly builds the population $F$ into the analysis. How can we adapt EB to directly model the population distribution of data?

### 2.2 POPULATION EMPIRICAL BAYES

We take a two-step approach. We first introduce an additional latent variable into EB and then use it to define a conditional density on $\boldsymbol{\theta}$.

The new variable $\mathbf{Z}$ is a latent dataset. This is a hypothetical dataset that we do not observe. It has the same size and dimension as the observed dataset $\mathbf{X}$, but with unknown observations. It lives one level above the $\boldsymbol{\theta}$ variables.

Given the latent dataset, we then define a conditional density on $\boldsymbol{\theta}$. We choose a density that depends on the latent dataset, $p(\boldsymbol{\theta} \mid \mathbf{Z})$. Figure 3 shows the framework.

This is a valid EB model; we simply propose a conditional density on $\boldsymbol{\theta}$ via a new latent variable $\mathbf{Z}$. However, it is incomplete. We must also choose a prior on $\mathbf{Z}$ and then define the form of $p(\boldsymbol{\theta} \mid \mathbf{Z})$.

We set the prior on the latent dataset $\mathbf{Z}$ to be the population $F$. The unknown, model-independent distribution of data acts as the prior on the latent dataset $\mathbf{Z}$.

We match the conditional density on $\boldsymbol{\theta}$ to the Bayesian posterior density of the original Bayesian model. Instead of conditioning on the observed dataset $\mathbf{X}$, we condition on the latent dataset $\mathbf{Z}$. This is how we interface the frequentist population distribution with the Bayesian model.

This fully describes the POP-EB framework. (We motivate these choices in Section 2.3.)

**Definition.** POP-EB defines the joint density

$$p(\boldsymbol{x}_{\text{new}}, \boldsymbol{\theta}_{\text{new}}, \mathbf{X}, \boldsymbol{\theta}, \mathbf{Z}) = p(\boldsymbol{x}_{\text{new}} \mid \boldsymbol{\theta}_{\text{new}}) \, p(\boldsymbol{\theta}_{\text{new}} \mid \mathbf{Z})$$
$$\times \prod_{n=1}^{N} p(\boldsymbol{x}_n \mid \boldsymbol{\theta}_n) \, p(\boldsymbol{\theta}_n \mid \mathbf{Z})$$
$$\times F(\mathbf{Z}). \qquad (3)$$

To obtain a predictive density from the joint density, we follow the same recipe as before: condition on the observations $\mathbf{X}$ and marginalize over the latent variables.

Marginalizing over $\boldsymbol{\theta}$ is straightforward. In addition, we marginalize over the latent dataset $\mathbf{Z}$, which gives the predictive density

$$p(\boldsymbol{x}_{\text{new}} \mid \mathbf{X}) = \int p(\boldsymbol{x}_{\text{new}} \mid \mathbf{Z}) \, p(\mathbf{Z} \mid \mathbf{X}) \, \mathrm{d}\mathbf{Z}. \qquad (4)$$

This is the POP-EB predictive density. It integrates the Bayesian predictive density over the latent dataset $\mathbf{Z}$ using the conditional density

$$p(\mathbf{Z} \mid \mathbf{X}) = \frac{p(\mathbf{X} \mid \mathbf{Z}) \, F(\mathbf{Z})}{\int p(\mathbf{X} \mid \mathbf{Z}') \, F(\mathbf{Z}') \, \mathrm{d}\mathbf{Z}'}. \qquad (5)$$

This is a key conditional density in POP-EB. It describes how the latent dataset varies given the observed dataset. The original Bayesian model prescribes the form of $p(\mathbf{X} \mid \mathbf{Z})$ and the population $F(\mathbf{Z})$ factors in as the prior.

Thus, POP-EB directly incorporates the population $F$ as a prior on the latent dataset. But the population is, by definition, unknown. We address this next.

**Plug-in principle.** The empirical population $\widehat{F}$ is the distribution that puts weight $1/N$ on each observation in the observed dataset $\{\boldsymbol{x}_n\}_1^N$. The plug-in estimator of a function of $F$ is simply the same function evaluated with $\widehat{F}$ instead. In the absence of any other information about the population, the plug-in principle is asymptotically efficient (Efron and Tibshirani, 1994).

The plug-in principle provides a nonparametric estimate of $F$. It enjoys a tight connection to the bootstrap and related techniques. We discuss computation in greater detail in Section 3. But now, we owe the reader an explanation.

## 2.3 MOTIVATION

Here is the story so far. The population $F$ is the model-independent mechanism that generates $\mathbf{X}$. Bayesian analysis employs a model $p(\mathbf{X}, \boldsymbol{\theta})$. The model is helpful; it gives strength to the statistical analysis (Young and Smith, 2005). However, Bayesian theory assumes the model is correct.[2] This is not always true; the model is often misspecified.

We focus on predictive inference. Our goal is to help a Bayesian model provide the best predictive accuracy, in spite of model misspecification. So, we seek a way to incorporate the model-independent population $F$ into our model-based Bayesian analysis.

This is why we set $F$ as the prior on the latent dataset $\mathbf{Z}$. In principle, there is no obstacle in using any other prior density. The prior on $\mathbf{Z}$ captures knowledge about the data generating mechanism that might otherwise be difficult to express in the model. In our case, this knowledge is precisely the population distribution $F$.

Consistency motivates our design of the conditional density on $\boldsymbol{\theta}$. Any density that depends on $\mathbf{Z}$ would be valid. We choose the Bayesian posterior density to mimic the original Bayesian model. Consider the Bayesian predictive density from Equation (1). It integrates the likelihood over the posterior density of the latent variables. The POP-EB predictive density of Equation (4) mirrors this form by integrating the Bayesian predictive density over the posterior density of the latent dataset.

## 3 COMPUTATION

We describe two empirical approximations to the POP-EB predictive density, develop some insight through simulation, and study a linear regression model with real data. More results follow in the empirical study of Section 5.

### 3.1 BOOTSTRAP

The plug-in principle replaces the population $F$ with its empirical counterpart $\widehat{F}$. However, direct computation with $\widehat{F}$ is also challenging. The POP-EB predictive density requires evaluating $p(\mathbf{Z} \mid \mathbf{X})$. This involves considering all possible datasets in the support of $\widehat{F}$, an intractable task.

The bootstrap is a computational technique for approximating functions of $\widehat{F}$. Define the bootstrapped dataset as $\mathbf{Z}^{(b)} = \{x_n^{(b)}\}_1^N$ where each $x_n^{(b)}$ is uniformly sampled from $\widehat{F}$ *with* replacement. The bootstrapped dataset $\mathbf{Z}^{(b)}$ contains as many observations as $\mathbf{X}$; some observations appear multiple times, others not at all.

All boostrapped datasets are equally likely. So we can approximate calculations of $\widehat{F}$ with a uniform distribution.

Call this distribution $\widehat{G}$; it is a discrete uniform distribution over the $B$ bootstrapped datasets $\{\mathbf{Z}^{(b)}\}_1^B$. This reduces the space of possible datasets to $\mathcal{O}(B)$.

### 3.2 MAP APPROXIMATION

A simple way to approximate the POP-EB predictive density is maximum a posteriori (MAP) estimation. In our setup, we find the most likely latent dataset $\mathbf{Z}^*$ from $p(\mathbf{Z} \mid \mathbf{X})$ and plug it into the Bayesian predictive density.

The POP-EB MAP predictive density is simply

$$p_{\text{MAP}}(x_{\text{new}} \mid \mathbf{X}) = p(x_{\text{new}} \mid \mathbf{Z}^*),$$

where the most likely dataset is

$$\begin{aligned}\mathbf{Z}^* &= \arg\max_{\mathbf{Z}} p(\mathbf{Z} \mid \mathbf{X}) \\ &= \arg\max_{\mathbf{Z}} p(\mathbf{X} \mid \mathbf{Z}) \, \widehat{F}(\mathbf{Z}).\end{aligned}$$

MAP estimation evades the intractable denominator in Equation (5), as the maximization only depends on $\mathbf{Z}$. However, the maximization is still intractable, so we replace $\widehat{F}$ with $\widehat{G}$ to get

$$\mathbf{Z}^{(b^*)} \approx \arg\max_{\mathbf{Z}^{(b)} \in \widehat{G}} p(\mathbf{X} \mid \mathbf{Z}^{(b)}).$$

This turns out to be a special case of the bumping technique (Tibshirani and Knight, 1999). Bumping is a bootstrap-based method to fit arbitrary models to a dataset; it finds the fit that minimizes some metric over bootstrapped datasets. Our metric is the average predictive density evaluated on the original dataset. Evaluating the metric on the original dataset guards against overfitting; it includes both held-in and held-out samples.

Enumeration spells the procedure out:

1. Draw $B$ bootstrapped datasets. This gives a set of datasets $\{\mathbf{Z}^{(b)}\}_1^B \sim \widehat{G}$.[3]

2. For each bootstrap index $b = 1, \cdots, B$:
   Compute and evaluate the Bayesian predictive density on the original dataset
   
   $$p(\mathbf{X} \mid \mathbf{Z}^{(b)}) = \prod_{n=1}^N p(x_n \mid \mathbf{Z}^{(b)}).$$

3. Return the bootstrap index $b^*$ that maximizes the Bayesian predictive density above.

We approximate the POP-EB MAP predictive density as

$$p_{\text{MAP}}(x_{\text{new}} \mid \mathbf{X}) \approx \int p(x_{\text{new}} \mid \boldsymbol{\theta}_{\text{new}}) \, p(\boldsymbol{\theta}_{\text{new}} \mid \mathbf{Z}^{(b^*)}) \, d\boldsymbol{\theta}_{\text{new}}. \tag{6}$$

---

[2] Bernardo and Smith (2000) identify this as the $\mathcal{M}$-closed view.

[3] Tibshirani and Knight (1999) recommend including the observed dataset. We follow their advice.

It has the same form as the Bayesian predictive density, but integrates over the Bayesian posterior conditioned on the bootstrapped dataset $\mathbf{Z}^{(b^*)}$.

## 3.3 FULL BAYESIAN APPROXIMATION

We also consider directly evaluating the POP-EB predictive density of Equation (4). In general, the posterior $p(\mathbf{Z} \mid \mathbf{X})$ is intractable. Luckily, replacing $\widehat{F}$ with $\widehat{G}$ reduces the integral to a finite sum over $B$. We call this a full Bayes (FB) approximation, as it is an exact evaluation of the POP-EB predictive density under the $\widehat{G}$ approximation of the empirical population. (The supplement contains a derivation.)

The POP-EB FB predictive density is a weighted sum,

$$p_{\text{FB}}(x_{\text{new}} \mid \mathbf{X}) = \sum_{b=1}^{B} w_b \, p(x_{\text{new}} \mid \mathbf{Z}^{(b)}),$$

where the weights are

$$w_b = \frac{p(\mathbf{X} \mid \mathbf{Z}^{(b)}) \, \widehat{G}(\mathbf{Z}^{(b)})}{\sum_b p(\mathbf{X} \mid \mathbf{Z}^{(b)}) \, \widehat{G}(\mathbf{Z}^{(b)})} = \frac{p(\mathbf{X} \mid \mathbf{Z}^{(b)})}{\sum_b p(\mathbf{X} \mid \mathbf{Z}^{(b)})},$$

and the $\mathbf{Z}^{(b)}$ and drawn from $\widehat{G}$. The probabilities $\widehat{G}(\mathbf{Z}^{(b)})$ are all equal to $1/B$, and so they disappear. As with the MAP case, the intractable denominator from Equation (5) also drops out of the calculation.

## 3.4 SIMULATION STUDY

To investigate these POP-EB quantities, we construct a toy example of model mismatch. The example is intentionally simple; it provides insight into why and how POP-EB mitigates model misspecification.

Consider that we typically observe data from a Poisson distribution with rate $\theta = 5$. (For instance, a router receives packets over a network; the wait-times are the measurements.) We model the data using a Poisson likelihood $p(x \mid \theta) = \text{Poisson}(\theta)$, and center a Gamma prior at $\theta = 5$ as $p(\theta) = \text{Gam}(\alpha = 2.5, \, \beta = 0.5)$.

Imagine that five percent of the time, the network fails. During these failures, the wait-times come from a different Poisson with rate $\theta = 50$. The population describes a mixture of two Poisson distributions. But the single Poisson model likelihood does not. A good predictive density should accurately describe the population.

Figure 4a shows the predictive densities. The Bayesian predictive density exhibits poor predictive accuracy; it describes neither of the two Poisson distributions in the population. In contrast, both POP-EB predictive densities match the dominant Poisson distribution. POP-EB uses the empirical population to focus on the observations in the dataset that give it greater predictive power.

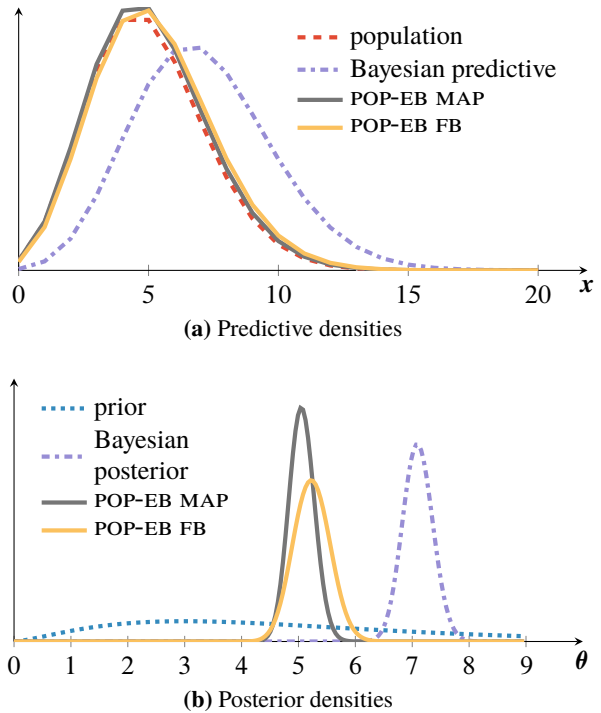

**Figure 4:** Gamma-Poisson simulation. The population in subpanel **(a)** has an small extra bump at 50 (not shown).

Figure 4b shows the underlying posterior densities on the latent variable $\theta$ that describes the wait-time. For POP-EB MAP this is the posterior density $p(\theta \mid \mathbf{Z}^{(b^*)})$; a single Gamma distribution that, when put through the Bayesian predictive density, gives the highest accuracy. For the POP-EB FB case, the posterior is a weighted sum of Gamma distributions. Both POP-EB posterior densities sit closer to the dominant rate of $\theta = 5$. This explains the behavior of the POP-EB predictive densities in Figure 4a.

The POP-EB MAP predictive density is easy to compute; it is a negative binomial distribution, like the Bayesian predictive density. In contrast, the POP-EB FB predictive density must be numerically calculated using all bootstrap samples; it has no simple analytic form.

We generate these plots using $B = 100$ bootstrapped datasets. The exact shape of the POP-EB densities depend on the bootstrapped datasets, but they are reproducibly closer to $\theta = 5$. The results are also insensitive to different prior configurations, such as a sharp prior centered at $\theta = 5$. EB is also of little use here; it estimates a nearly flat prior on $\theta$ from the data. (See supplementary note and code.)

## 3.5 BAYESIAN LINEAR REGRESSION

We now apply the POP-EB framework to a real-world dataset. Consider a Bayesian linear regression model. The likelihood is a Gaussian distribution and the latent random variables are the regression coefficients $\boldsymbol{\beta}$ and the variance $\sigma^2$. Using conjugate priors (Gaussian for the coefficients and inverse Gamma for the variance) gives a Bayesian predictive density that follows a t-distribution (Murphy, 2012). We posit an uninformative prior.

**Table 1:** Bayesian linear regression predictive accuracy results on random held-out splits of the `bodyfat` dataset.

|  | Split #1 | | | Split #2 | | |
| --- | --- | --- | --- | --- | --- | --- |
|  | logp | MSE | MAE | logp | MSE | MAE |
| Bayes | 0.67 | 5.2e-3 | 5.7e-2 | 0.83 | 7.6e-3 | 6.6e-2 |
| POP-EB MAP | **1.26** | 3.2e-3 | 4.3e-2 | 1.18 | 5.1e-3 | 5.4e-2 |
| POP-EB FB | 1.25 | **3.0e-3** | **4.2e-2** | 1.18 | **4.4e-3** | **5.2e-2** |
|  | Split #3 | | | Split #4 | | |
|  | logp | MSE | MAE | logp | MSE | MAE |
| Bayes | 0.85 | 6.5e-3 | 5.4e-2 | 0.81 | 7.4e-3 | 6.7e-2 |
| POP-EB MAP | **1.24** | 4.0e-3 | 4.9e-2 | **1.21** | 5.1e-3 | 5.5e-2 |
| POP-EB FB | 1.23 | **3.3e-3** | **4.4e-2** | 1.19 | **4.2e-3** | **5.3e-2** |
|  | Split #5 | | | Split #6 | | |
|  | logp | MSE | MAE | logp | MSE | MAE |
| Bayes | 0.82 | 7.2e-3 | 6.9e-2 | 0.44 | 3.1e-2 | 9.2e-2 |
| POP-EB MAP | **1.15** | 6.2e-3 | 6.1e-2 | 0.75 | 1.7e-2 | 7.1e-2 |
| POP-EB FB | 1.14 | **4.9e-3** | **5.2e-2** | **0.76** | **1.6e-2** | **6.6e-2** |

We study the predictive performance of both POP-EB predictive densities on the `bodyfat` dataset (StatLib, 1995). Accurate measurements of body fat are costly. The dataset contains the body fat percentages of $N = 252$ men along with 14 other features that are cheaper to measure. We want to predict body fat percentage using these 14 features.

We randomly extract datasets of 200 measurements and hold the remaining 52 to evaluate predictive accuracy. Table 1 reports the average logarithm of the predictive densities (logp), along with mean squared error (MSE) and mean absolute error (MAE). In all cases, the POP-EB predictive densities reach higher predictive accuracy on held-out data. The POP-EB FB density performs similarly to the POP-EB MAP density, but exhibits slightly better MSE and MAE values. Since the dataset is small, we split it six times. These results are reproducible across a variety of splits. We use $B = 25$ bootstrap samples. (See supplementary code.)

The POP-EB FB density is a better approximation to the POP-EB predictive density than its MAP counterpart. The results in Table 1 corroborate this. However, POP-EB FB density lack an analytic form. In contrast, the POP-EB MAP predictive density offers an attractive improvement in predictive accuracy while maintaining the form of the Bayesian predictive density. We now turn to approximating it for complex Bayesian models.

## 4 POPULATION EMPIRICAL VARIATIONAL INFERENCE

Modern Bayesian statistics and machine learning has moved well past simple conjugate models like Bayesian linear regression. In complex models, such as Bayesian mixture models (Bishop, 2006) or probabilistic topic models (Blei et al., 2003), the posterior and predictive densities are intractable to compute. Monte Carlo sampling and variational techniques are two popular frameworks for approximation.

In this section, we develop an efficient variational approximation to the POP-EB MAP predictive density.

### 4.1 VARIATIONAL INFERENCE

Variational inference is an optimization-based approach to approximate posterior computation in a Bayesian model $p(\mathbf{X}, \boldsymbol{\theta})$ (Jordan et al., 1999; Wainwright and Jordan, 2008). The idea is to posit a simple parameterized density family over the latent variables, $q(\boldsymbol{\theta} ; \boldsymbol{\lambda})$. We then seek the member of the family that is closest in Kullback-Leibler (KL) divergence to the true posterior density $p(\boldsymbol{\theta} \mid \mathbf{X})$. Minimizing KL $(q \parallel p)$ is equivalent to maximizing a lower bound on the marginal density of the data. This gives a variational objective function, called the evidence lower bound (ELBO)

$$\mathcal{L}(\mathbf{X}, \boldsymbol{\lambda}) = \mathbb{E}_q [\log p(\mathbf{X}, \boldsymbol{\theta})] - \mathbb{E}_q [\log q(\boldsymbol{\theta} ; \boldsymbol{\lambda})].$$

Variational inference maximizes this objective function with respect to the set of parameters $\boldsymbol{\lambda}$.

In mean-field variational inference, we assume the variational density fully factorizes. This divides the variational parameters into $M$ parts, $\boldsymbol{\lambda} = \{\boldsymbol{\lambda}_m\}_1^M$. We then maximize the ELBO using coordinate ascent (Jordan et al., 1999). This means iteratively maximizing one variational parameter at a time, holding all others fixed. The separation of latent variables leads to independent updates for each variational parameter. This coordinate ascent scheme guarantees convergence to a local maximum of the ELBO (Bishop, 2006).

### 4.2 APPROXIMATING THE POP-EB MAP PREDICTIVE DENSITY

Our aim is to employ variational inference to approximate the POP-EB MAP predictive density. It shares the same form as the Bayesian predictive density, but integrates over the Bayesian posterior density evaluated on a particular bootstrapped dataset. To that end, consider the ELBO evaluated on a bootstrapped dataset, $\mathcal{L}(\mathbf{Z}^{(b)}, \boldsymbol{\lambda})$.

The variational objective becomes a joint optimization of the variational parameters and the bootstrap index

$$\boldsymbol{\lambda}_\dagger^{(b^*)} = \arg\max_{\boldsymbol{\lambda}} \mathcal{L}(\mathbf{Z}^{(b^*)}, \boldsymbol{\lambda})$$

$$b^* = \arg\max_b \prod_{n=1}^N \int p(x_n \mid \boldsymbol{\theta}_n) q\left(\boldsymbol{\theta}_n ; \boldsymbol{\lambda}_\dagger^{(b)}\right) d\boldsymbol{\theta}_n.$$

The two optimization problems are coupled: the first seeks the best approximation to the Bayesian posterior density while the second seeks the latent dataset index that gives the highest predictive accuracy. The variational density $q(\boldsymbol{\theta} \,;\, \boldsymbol{\lambda}_{\dagger}^{(b^*)})$ is the closest KL approximation to the Bayesian posterior density inside the POP-EB MAP predictive density of Equation (6).

The naïve way to solve the joint optimization above is to adapt our earlier technique from Section 3.2. Bootstrap the dataset $B$ times, maximize the ELBO for each dataset, and choose the one that gives the best predictive performance. This is a costly procedure. It requires multiple maximizations of the ELBO, which we wish to avoid.

### 4.3 BUMPING VARIATIONAL INFERENCE

We propose a stochastic optimization algorithm that maximizes the ELBO once. We weave bumping into each iteration of the optimization. We call this method bumping variational inference (BUMP-VI) (Algorithm 1). At a high level, the algorithm works as follows.

Consider a single iteration. We bootstrap the dataset $B$ times. For each bootstrapped dataset, we compute a gradient $\boldsymbol{g}_{\boldsymbol{\lambda}}^{(b)}$ in the parameters $\boldsymbol{\lambda}$. These gradients are noisy estimates of the "true" gradient, had we taken the naïve approach above and determined which bootstrapped dataset $\mathbf{Z}^{(b^*)}$ led to the best predictive performance. The cost of computing $B$ gradients is small in many Bayesian models. (The supplement describes an efficient implementation.)

Then, we employ the bumping procedure from Section 3.2. This means taking a step in the parameters for each bootstrapped dataset and evaluating the Bayesian predictive density on the original dataset. This evaluation is the main added cost of BUMP-VI over classical variational methods. We pick the index $b^{(*)}$ that gives the highest predictive performance and take a step in the direction it indexes.

BUMP-VI is a stochastic optimization method; the step-size sequence matters for establishing convergence guarantees (Robbins and Monro, 1951). We use a constant step-size as it isolates the performance of the algorithm from any sequence-related effects and provides quantifiable error bounds (Nemirovski et al., 2009). Code is available at https://github.com/Blei-Lab/lda-bump-cpp.

## 5 EMPIRICAL STUDY

We apply BUMP-VI to two complex tasks: Bayesian mixture modeling of image histograms and latent Dirichlet allocation (LDA) topic modeling of a scientific text corpus. We compare BUMP-VI to coordinate ascent variational inference. In both examples, BUMP-VI uniformly reaches higher predictive accuracy. We first present both sets of results and then discuss performance.

---

**Algorithm 1:** Bumping Variational Inference

**Input**: Model $p(\mathbf{X}, \boldsymbol{\theta})$, variational family $q(\boldsymbol{\theta} \,;\, \boldsymbol{\lambda})$
**Result**: Optimized $q(\boldsymbol{\theta} \,;\, \boldsymbol{\lambda}_{\dagger}^{(b^*)})$ approximation

Choose the number of bootstraps $B$
Choose a step-size sequence $\{\rho_{(i)}\}_1^\infty$
Initialize parameters $\boldsymbol{\lambda}_{(0)}$, iteration $i = 0$

**while** $\|\boldsymbol{\lambda}_{(i+1)} - \boldsymbol{\lambda}_{(i)}\|$ *is above some threshold* **do**
  **for** $b = 1$ **to** $B$ **do**
    Draw a bootstrap sample $\mathbf{Z}^{(b)} \sim \hat{G}$
    Calculate gradient of parameters $\boldsymbol{\lambda}$
    $$\boldsymbol{g}_{\boldsymbol{\lambda}}^{(b)} \longleftarrow \nabla_{\boldsymbol{\lambda}} \mathcal{L}(\mathbf{Z}^{(b)}, \boldsymbol{\lambda})$$
  **end**
  Choose the bootstrap index that maximizes the Bayesian predictive density on the dataset
  $$b^* \longleftarrow \arg\max_b \prod_{n=1}^N \int p(\boldsymbol{x}_n \mid \boldsymbol{\theta}_n) \\ q(\boldsymbol{\theta}_n \,;\, \boldsymbol{\lambda}_{(i)} + \rho_{(i)} \boldsymbol{g}_{\boldsymbol{\lambda}}^{(b)}) \\ \mathrm{d}\boldsymbol{\theta}_n$$
  Take a step in direction indexed by $b^*$
  $$\boldsymbol{\lambda}_{(i+1)} \longleftarrow \boldsymbol{\lambda}_{(i)} + \rho_{(i)} \boldsymbol{g}_{\boldsymbol{\lambda}}^{(b^*)}$$
  Update iteration counter
  $$i \longleftarrow i + 1$$
**end**
Return variational parameters
$$\boldsymbol{\lambda}_{\dagger}^{(b^*)} \longleftarrow \boldsymbol{\lambda}_{(i+1)}$$

---

### 5.1 BAYESIAN MIXTURE MODEL

Consider a Gaussian mixture model (GMM) with a Gaussian-Gamma prior on the mixture component means and precisions, and a Dirichlet prior on the proportions. These are conditionally conjugate priors that lead to straightforward coordinate ascent update equations (Bishop, 2006).

The imageCLEF dataset has 576-dimensional color histograms of natural images (Villegas et al., 2013). We randomly select 5 000 images as our dataset and choose another 1 000 images for predictive accuracy tests. We standardize the mean and variance of the set of histograms. Thus, the hyperparameters for the Gaussian-Gamma prior are zero on the component means and one on the precisions. The hyperparameter on the Dirichlet is $1/K$ where $K$ is the number of components. We set the number of bootstrap samples to $B = 10$ and the step-size to $\rho = 0.1$.

We compare coordinate ascent to BUMP-VI across a range of components. Figure 5 displays these results. BUMP-VI attains a higher average log predictive evaluated on the held-out set. We run each algorithm ten times per configuration to account for random initialization of the initial parameters.

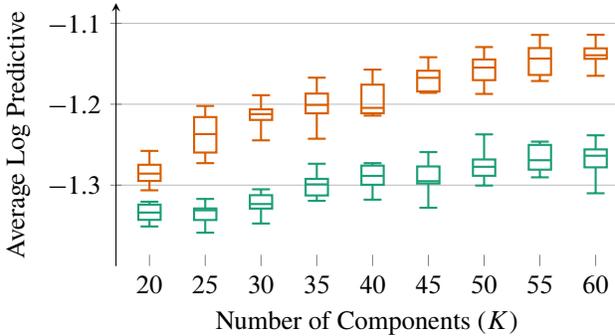

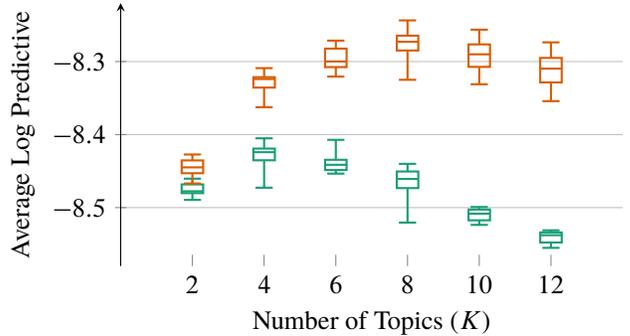

**Figure 5:** Average log-predictive density of 1 000 held-out images. BUMP-VI (orange) reaches a higher predictive accuracy than coordinate ascent (green) across a range of configurations. Boxplots show the median, quartiles, and 1.5 interquartile range of ten repeats per configuration.

**Figure 7:** Average per-word log-predictive density of 1 000 held-out abstracts. BUMP-VI (orange) reaches a higher predictive accuracy than coordinate ascent (green) across a range of model configurations.

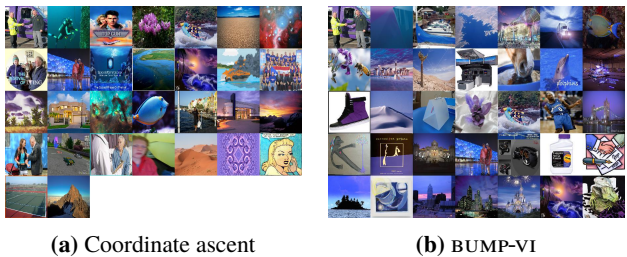

**(a)** Coordinate ascent    **(b)** BUMP-VI

**Figure 6:** The "blue-purple" GMM ($K = 40$) coordinate ascent component has some extraneous images. The bumping component is qualitatively more uniform.

Both algorithms converge in fewer than 75 iterations.

Figure 6 presents the effect of reaching higher predictive accuracy. We assign each image to its most probable mixture component and pick the one with "blue-purple" images. Though this is a subjective matter, the BUMP-VI component appears more uniform than its counterpart. The coordinate ascent component seems to have red-toned images that might belong in a different component.

### 5.2 LATENT DIRICHLET ALLOCATION

We study LDA with a corpus of 5 000 randomly selected abstracts from the arXiv repository. The abstracts are short (~150 words) and the vocabulary is large (~12 000 unique words). The arXiv exhibits jargon-heavy abstracts, which leads to a large vocabulary. This makes for a challenging dataset; because of the vocabulary size, we only expect to identify a few topics in a corpus of size 5 000.

LDA has two Dirichlet priors. We set the hyperparameter on the topics distributions to be $1/K$ where $K$ is the number of topics. The hyperparameter on the topics is 0.005, which is approximately $60/V$ where $V$ is the vocabulary size. We set the number of bootstrap samples to $B = 10$ and the step-size to $\rho = 0.05$.

We study BUMP-VI across a range of topics. Figure 7 plots a comparison to coordinate ascent. BUMP-VI attains a higher average per-word log predictive evaluated on the held-out set. Tables 3 and 4 show how this difference affects the topics. While six of the topics are similar, BUMP-VI finds two topics with higher predictive power. As with mixture components, examining topics is a subjective activity. Both algorithms converged in fewer than 150 iterations.

### 5.3 MITIGATING MODEL MISMATCH

There is no reason to believe that natural images are truly distributed as a mixture of Gaussians. The same holds for the "bag of words" assumption that underlies LDA. With real data, there is always some level of inherent model mismatch. BUMP-VI attempts to mitigates this effect.

The performance gap widens with the number of components. We might expect a simpler model to be more severely mismatched. This demonstrates that POP-EB cannot rectify the choice of an inappropriate model. For example, two topics are simply too few to accurately model our arXiv corpus; both algorithms do poorly. However, with twelve topics, coordinate ascent does even worse. This may be due to other effects, such as local maxima in the variational objective function (Wainwright and Jordan, 2008).

Does BUMP-VI outperform classical techniques because it reaches a better local maximum of the ELBO? The answer is no. Evaluating the ELBO on the observed dataset gives similar numerical values for both methods. (In our experiments, coordinate ascent usually reaches a slightly higher number than BUMP-VI.) This is not surprising, as BUMP-VI solves a different optimization problem than coordinate ascent variational inference; it approximates the POP-EB MAP predictive density instead of the Bayesian predictive density.

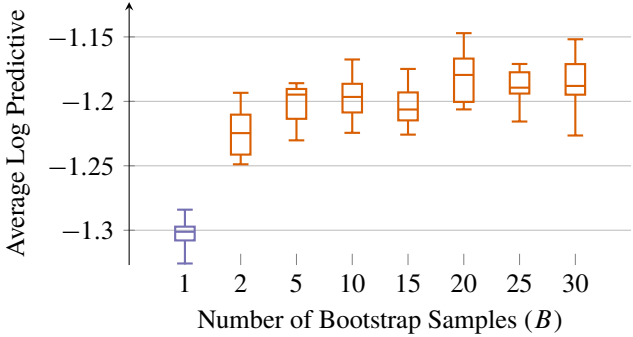

**Figure 8:** BUMP-VI sensitivity study with the GMM ($K = 40$) model. Average log-predictive density of 1 000 held-out images, over ten repeats.

**Table 2:** Runtime ratios of BUMP-VI compared to coordinate ascent. An efficient re-weighting strategy gives ratios that are less than $B$. (See supplementary note.)

| $B$ | 2 | 5 | 10 | 15 | 30 |
|---|---|---|---|---|---|
| BUMP-VI | 1.8× | 3.2× | 5.6× | 7.7× | 15.1× |

### 5.4 SENSITIVITY TO BOOTSTRAP SAMPLES AND COMPUTATIONAL COST

We also study the sensitivity of BUMP-VI to the number of bootstrap samples $B$ (Figure 8). Tibshirani and Knight (1999) recommend using 20-30 bootstrap samples for bumping, but as few as five appear to perform well. This is because BUMP-VI, in a loose sense, re-samples the dataset $B$ times the number of iterations. For completeness, we also compare the case of $B = 1$. This is equivalent to stochastic variational inference (SVI) with "minibatch" size equal to $N$ (Hoffman et al., 2013). This shows that the performance gain in BUMP-VI is not due to stochastic escapes from local maxima. (Though, it likely benefits from it.)

The POP-EB framework incurs a computational cost proportional to the choice of $B$. BUMP-VI reduces this cost in two ways. The first is due to the iterative resampling effect mentioned above. The second is due to an efficient calculation of the $B$ gradients at each iteration. (See supplementary note.) Table 2 shows that BUMP-VI is less than $B$ times the cost of coordinate ascent variational inference.

### 6 CONCLUSION

Mismatched models exhibit poor predictive performance. The POP-EB predictive density mitigates this effect by incorporating the population $F$ into Bayesian analysis. The POP-EB MAP predictive density is attractively simple; it explores the space of bootstrapped dataset to find the one with highest predictive power. BUMP-VI extends this idea to variational inference; it delivers an efficient algorithm with promising results on real-world datasets.

POP-EB, like EB, uses the dataset twice: once to estimate the prior and again during inference. There are alternatives to consider, such as cross-validation and bias correction (Efron and Tibshirani, 1997; Gelman et al., 2013). Directly modeling the expected predictive performance of future data should also improve POP-EB.

**Table 3:** Coordinate ascent LDA topics ($K = 8$).

| Topic 1 | Topic 2 | Topic 3 | Topic 4 |
|---|---|---|---|
| charge | gravitational | et | knowledge |
| ground | dynamical | al | often |
| induced | cosmological | accurate | introduced |
| regime | physics | test | novel |
| length | review | extended | among |
| leads | black | identified | research |
| dependent | background | signal | called |
| transport | universe | period | easily |
| fluctuations | gr | during | key |
| scattering | gravity | correlation | processing |

| Topic 5 | Topic 6 | Topic 7 | Topic 8 |
|---|---|---|---|
| invariant | quark | algorithm | galaxies |
| algebra | cross | random | stars |
| operator | production | applications | galaxy |
| theorem | heavy | efficient | stellar |
| defined | qcd | finally | gas |
| g | collisions | probability | galactic |
| generalized | predictions | network | sources |
| complex | corrections | optimal | objects |
| explicit | experiment | gaussian | source |
| infinite | photon | distributed | sample |

**Table 4:** BUMP-VI LDA topics ($K = 8$).

| Topic 1 | Topic 2 | Topic 3 | Topic 4 |
|---|---|---|---|
| induced | cosmological | accurate | performance |
| charge | universe | test | research |
| regime | gravitational | identified | development |
| electronic | review | during | novel |
| transport | cosmic | better | key |
| length | dark | mostly | developed |
| leads | gravity | errors | processing |
| temperatures | background | scales | called |
| frequency | relativity | though | knowledge |
| influence | physics | sensitivity | analyze |

| Topic 5 | Topic 6 | Topic 7 | Topic 8 |
|---|---|---|---|
| algebra | quark | algorithm | galaxies |
| invariant | production | random | stars |
| defined | cross | probability | galaxy |
| operator | heavy | network | stellar |
| theorem | qcd | applications | galactic |
| g | collisions | quant | gas |
| complex | predictions | complexity | sample |
| construct | momentum | finally | sources |
| conjecture | photon | problems | objects |
| special | detector | optimal | source |

**Acknowledgments**

We thank Allison Chaney, Laurent Charlin, Stephan Mandt, Kui Tang, Yixin Wang, and the reviewers for their insightful comments. DMB is supported by NSF IIS-1247664, ONR N00014-11-1-0651, and DARPA FA8750-14-2-0009.

# SUPPLEMENTARY MATERIAL

## THE POP-EB FULL BAYES APPROXIMATION

Recall the form of the population empirical Bayes (POP-EB) predictive density,

$$p(x_{\text{new}} \mid \mathbf{X}) = \int p(x_{\text{new}} \mid \mathbf{Z}) \, p(\mathbf{Z} \mid \mathbf{X}) \, d\mathbf{Z}.$$

Expanding the conditional density at the end gives,

$$p(x_{\text{new}} \mid \mathbf{X}) = \int p(x_{\text{new}} \mid \mathbf{Z}) \frac{p(\mathbf{X} \mid \mathbf{Z}) \, F(\mathbf{Z})}{\int p(\mathbf{X} \mid \mathbf{Z}') \, F(\mathbf{Z}') \, d\mathbf{Z}'} d\mathbf{Z}.$$

The plug-in principle replaces $F$ with the empirical distribution of data $\widehat{F}$. We then approximate the empirical distribution using the bootstrap by replacing $\widehat{F}$ with $\widehat{G}$. This leads to the approximation

$$p(x_{\text{new}} \mid \mathbf{X}) \approx \sum_{\mathbf{Z}} p(x_{\text{new}} \mid \mathbf{Z}) \frac{p(\mathbf{X} \mid \mathbf{Z}) \, \widehat{G}(\mathbf{Z})}{\sum_{\mathbf{Z}'} p(\mathbf{X} \mid \mathbf{Z}') \, \widehat{G}(\mathbf{Z}')},$$

because $\widehat{G}$ is a discrete distribution.

Specifically, $\widehat{G}$ is uniform over a set of $b = 1, \ldots, B$ bootstrapped datasets, which gives

$$p(x_{\text{new}} \mid \mathbf{X}) \approx \sum_{b=1}^{B} p(x_{\text{new}} \mid \mathbf{Z}^{(b)}) \frac{p(\mathbf{X} \mid \mathbf{Z}^{(b)}) \frac{1}{B}}{\sum_{b=1}^{B} p(\mathbf{X} \mid \mathbf{Z}^{(b)}) \frac{1}{B}}.$$

The POP-EB full Bayes (FB) is then

$$p_{\text{FB}}(x_{\text{new}} \mid \mathbf{X}) = \sum_{b=1}^{B} w_b \, p(x_{\text{new}} \mid \mathbf{Z}^{(i)}),$$

with weights

$$w_b = \frac{p(\mathbf{X} \mid \mathbf{Z}^{(b)})}{\sum_{b=1}^{B} p(\mathbf{X} \mid \mathbf{Z}^{(b)})}.$$

## SIMULATION RESULTS WITH A SHARP PRIOR

Consider the model from the toy example in the paper. If we truly believe network failures are rare, we could posit an informative prior density. However, this has little effect in addressing model mismatch. Figure 1 shows the results of the same study under a very sharp prior centered at 5, $p(\theta) = \text{Gam}(\alpha = 500, \beta = 100)$.

The Bayesian posterior is more accurate than before; it shifts closer to the dominant rate of $\theta = 5$. This also moves the Bayesian predictive closer to the population. However, both POP-EB maximum a posteriori (MAP) and POP-EB full Bayes (FB) predictive densities still provide a better match to the true population.

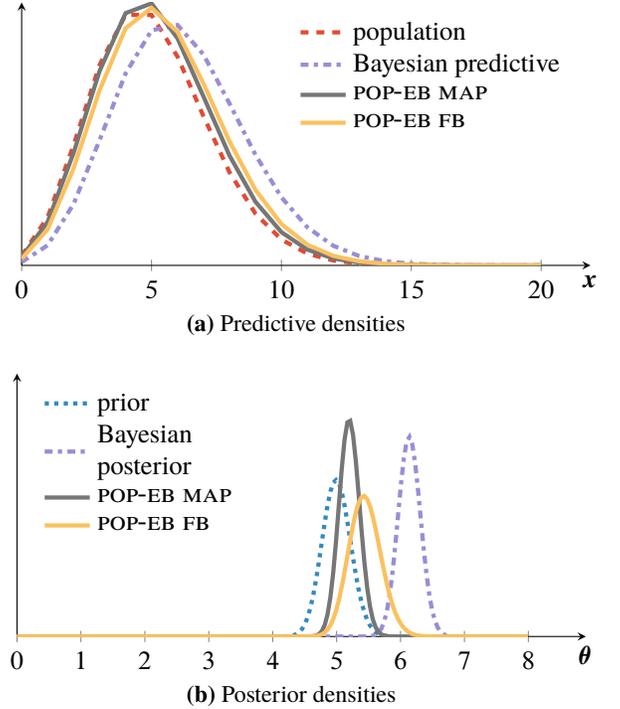

**Figure 1:** Gamma-Poisson model with a sharp prior density centered at 5. The population in subpanel **(a)** has an additional small bump at 50 (not shown).

## SIMULATION RESULTS WITH AN EMPIRICAL BAYES PRIOR

Empirical Bayes (EB) estimates the parameters of the prior density from the data. For simplicity, assume the prior is a Gamma distribution. One way to estimate the shape $\alpha$ and rate $\beta$ parameters is to match the mean and variance of the Gamma distribution to that of the data.

The Gamma distribution has mean $= \alpha/\beta$ and variance $= \alpha/\beta^2$. This leads to the following pair of equations

$$\frac{\alpha}{\beta} = \text{Mean}(\mathbf{X}) = \frac{1}{N} \sum_{n=1}^{N} x_n$$

$$\frac{\alpha}{\beta^2} = \text{Var}(\mathbf{X}) = \frac{1}{N} \sum_{n=1}^{N} (x_n - \text{Mean}(\mathbf{X}))^2.$$

The solution is

$$\alpha = \frac{[\text{Mean}(\mathbf{X})]^2}{\text{Var}(\mathbf{X})} \quad \text{and} \quad \beta = \frac{\text{Mean}(\mathbf{X})}{\text{Var}(\mathbf{X})}.$$

In our simulation study, these lead to estimates around $\alpha \approx 0.5$ and $\beta \approx 0.07$, which describes a nearly flat Gamma distribution. This does not help mitigate model mismatch, nor does it improve predictive accuracy.

# EFFICIENT BUMP-VI IMPLEMENTATION

We first modify our Bayesian model notation to mimic stochastic variational inference (SVI) (Hoffman et al., 2013). Consider a Bayesian model $p(\mathbf{X}, \boldsymbol{\theta})$. Separate the latent variables $\boldsymbol{\theta}$ into a set of local $\boldsymbol{\zeta}$ and global $\boldsymbol{\beta}$ variables. Local latent variables $\boldsymbol{\zeta} = \{\boldsymbol{\zeta}_n\}_1^N$ grow with the number of observations; global latent variables $\boldsymbol{\beta}$ do not. The likelihood becomes $p(X \mid \boldsymbol{\zeta}, \boldsymbol{\beta})$ and the prior $p(\boldsymbol{\zeta}, \boldsymbol{\beta})$.

Given the global variables $\boldsymbol{\beta}$, the local latent variable $\boldsymbol{\zeta}_n$, along with its observation $\boldsymbol{x}_n$, is conditionally independent of all other latent variables and observations

$$p(\boldsymbol{x}_n, \boldsymbol{\zeta}_n \mid \boldsymbol{x}_{-n}, \boldsymbol{\zeta}_{-n}, \boldsymbol{\beta}) = p(\boldsymbol{x}_n, \boldsymbol{\zeta}_n \mid \boldsymbol{\beta}).$$

The negative indexing notation means $\boldsymbol{x}_{-n} = \{x_i \mid i = 1, \ldots, n-1, n+1, \ldots, N\}$. Global latent variables lack such conditional independence.

This divide is natural in many models. For example, consider latent Dirichlet allocation (LDA). The global latent variables are the topics. (The number of topics is fixed and does not vary with the number of documents.) The local latent variables are the per-document topic distributions and the per-word assignments. (There are as many of these variables as documents and words within each document.)

The local-global separation simplifies the computation of the $B$ gradients in bumping variational inference (BUMP-VI). The variational family has two sets of variational parameters $q(\boldsymbol{\zeta}, \boldsymbol{\beta}\,; \boldsymbol{\phi}, \boldsymbol{\lambda})$ where $\boldsymbol{\phi}$ indexes the local variables and $\boldsymbol{\lambda}$ the global ones. This also splits the variational parameters in the evidence lower bound (ELBO) as $\mathcal{L}(\mathbf{X}, \boldsymbol{\phi}, \boldsymbol{\lambda})$.

Hoffman et al. (2013) show that the gradient calculation decomposes into a maximization of the local variables and a gradient with respect to the global variables. The recipe at each iteration is

$$\boldsymbol{\phi}_\dagger \leftarrow \arg\max_{\boldsymbol{\phi}} \mathcal{L}(\mathbf{X}, \boldsymbol{\phi}, \boldsymbol{\lambda}_{\text{prev}})$$
$$\boldsymbol{g}_{\boldsymbol{\lambda}} \leftarrow \nabla_{\boldsymbol{\lambda}} \mathcal{L}(\mathbf{X}, \boldsymbol{\phi}_\dagger, \boldsymbol{\lambda})$$
$$\boldsymbol{\lambda}_{\text{next}} \leftarrow \boldsymbol{\lambda}_{\text{prev}} + \rho \boldsymbol{g}_{\boldsymbol{\lambda}}.$$

where $\rho$ is a scalar step-size.

In SVI, we subsample the dataset $\mathbf{X}$ and accordingly re-weight the optimized local variables to construct an unbiased estimate of $\boldsymbol{g}_{\boldsymbol{\lambda}}$. Exponential family models parameterized in their natural forms enjoy a connection to their coordinate ascent updates, but the idea holds in general (Hoffman et al., 2013).

In BUMP-VI, we need $B$ gradients of the ELBO evaluated on the bootstrapped datasets $\{\mathbf{Z}^{(b)}\}_1^B$. Luckily, subsampling is conceptually equivalent to bootstrap resampling: they both induce a weighting scheme on the local latent variables.

We propose the following efficient implementation. At each iteration:

1. Compute the optimized local variables $\boldsymbol{\phi}_\dagger$ *once* for the original dataset.

2. Compute the form of $\nabla_{\boldsymbol{\lambda}} \mathcal{L}$.

3. Generate the $B$ gradients $\{\boldsymbol{g}_{\boldsymbol{\lambda}}^{(b)}\}_1^B$ by re-weighting the local variables according to the bootstrapped datasets $\{\mathbf{Z}^{(b)}\}_1^B$. This means weighting each local variable proportional to the number of times its paired observation appears in the bootstrapped dataset.

We implement this stragey in the accompany code: https://github.com/Blei-Lab/lda-bump-cpp.